\documentclass[12pt, draftclsnofoot, onecolumn]{IEEEtran}
\usepackage[utf8]{inputenc}

\usepackage{amsmath}
\usepackage{amssymb}
\usepackage{amsfonts}
\usepackage{cite}

\usepackage{booktabs}
\usepackage{bbm}
\usepackage{mathrsfs}

\DeclareMathOperator*{\argmax}{arg\,max}

\newcommand{\Rmnum}[1]{\expandafter\@slowromancap\romannumeral #1@}
\usepackage{graphicx}
\usepackage{subfigure}
\usepackage{epsfig}
\usepackage[numbers,sort&compress]{natbib}
\usepackage{xcolor}
\usepackage{color}
\usepackage{enumerate}
\usepackage{extarrows}
\usepackage[ruled,linesnumbered,lined]{algorithm2e}  
\usepackage{algorithmic}
\usepackage{tensor}
\usepackage[colorlinks,linkcolor=blue,anchorcolor=blue,citecolor=blue,urlcolor=black]{hyperref}
\usepackage{url}

\begin{document}

\title{Agreeing to Stop: Reliable Latency-Adaptive Decision Making via Ensembles of Spiking Neural Networks}
\author{Jiechen Chen, \IEEEmembership{Student Member,~IEEE}, Sangwoo Park,  \IEEEmembership{Member,~IEEE}, Osvaldo Simeone,~\IEEEmembership{Fellow,~IEEE}
\vspace{-0.5cm} 
\thanks{The authors are with the King’s Communications, Learning and Information Processing (KCLIP) lab, King’s College London, London, WC2R 2LS, UK. (email:\{jiechen.chen, sangwoo.park, osvaldo.simeone\}@kcl.ac.uk). This work was supported by the European Research Council (ERC) under the
European Union’s Horizon 2020 Research and Innovation Programme (grant agreement No. 725732), by the European Union’s Horizon Europe project CENTRIC (101096379), and by an Open Fellowship of the EPSRC (EP/W024101/1).}
}

\maketitle

\IEEEpeerreviewmaketitle
\newtheorem{definition}{\underline{Definition}}[section]
\newtheorem{fact}{Fact}
\newtheorem{assumption}{Assumption}
\newtheorem{theorem}{\underline{Theorem}}[section]
\newtheorem{lemma}{\underline{Lemma}}[section]
\newtheorem{proposition}{\underline{Proposition}}[section]
\newtheorem{corollary}[proposition]{\underline{Corollary}}
\newtheorem{example}{\underline{Example}}[section]
\newtheorem{remark}{\underline{Remark}}[section]
\newcommand{\mv}[1]{\mbox{\boldmath{$ #1 $}}}
\newcommand{\mb}[1]{\mathbb{#1}}
\newcommand{\Myfrac}[2]{\ensuremath{#1\mathord{\left/\right.\kern-\nulldelimiterspace}#2}}
\newcommand\Perms[2]{\tensor[^{#2}]P{_{#1}}}
\newcommand{\note}[1]{[\textcolor{red}{\textit{#1}}]}

\vspace{-0.2cm}

\begin{abstract}
Spiking neural networks (SNNs) are recurrent models that can leverage sparsity in input time series to efficiently carry out tasks such as classification. Additional efficiency gains can be obtained if decisions are taken as early as possible as a function of the complexity of the input time series. The decision on when to stop inference and produce a decision must rely on an estimate of the current accuracy of the decision. Prior work demonstrated the use of conformal prediction (CP) as a principled way to quantify uncertainty and support adaptive-latency decisions in SNNs. In this paper, we propose to enhance the uncertainty quantification capabilities of SNNs by implementing ensemble models for the purpose of improving the reliability of stopping decisions. Intuitively, an ensemble of multiple models can decide when to stop more reliably by selecting times at which most models agree that the current accuracy level is sufficient. The proposed method relies on different forms of information pooling from ensemble models, and offers theoretical reliability guarantees. We specifically show that variational inference-based ensembles with p-variable pooling significantly reduce the average latency of state-of-the-art methods, while maintaining reliability guarantees.
\end{abstract}

\section{Introduction} \label{intro}
\noindent \textbf{Context:} With the advent of large language models,  sequence models are currently among the most studied machine learning techniques. Unlike methods based on conventional neural networks, such as transformers, spiking neural networks (SNNs) process time series with the prime objective of optimizing energy efficiency, particularly in the presence of sparse inputs  \cite{8891810, ghosh2009spiking, tavanaei2019deep}. The energy consumption of an SNN depends on the number of spikes generated internally by the constituent spiking neurons \cite{mehonic2020memristors}, and inference energy can be further reduced if decisions are taken as early as possible as a function of the complexity of the input time series \cite{li2023unleashing}.   

The decision on when to stop inference and produce a decision must rely on an estimate of the current accuracy of the decision, as stopping too early may cause unacceptable drops in accuracy. 
The delay-adaptive rule proposed in \cite{li2023unleashing} uses the SNN's output confidence levels to estimate the true accuracy, while reference \cite{li2023seenn} determined the stopping time via a separate policy network.  SNN models, like their conventional neural network counterpart, tend to be poorly calibrated, producing overconfident decisions \cite{guo2017calibration} (see also Fig. 1 in \cite{chen2023spikecp}). As a consequence, the schemes in \cite{li2023unleashing,li2023seenn} do not offer any reliability guarantee at the stopping time. To address this problem, recent work \cite{chen2023spikecp} demonstrated the use of \emph{conformal prediction} (CP) \cite{angelopoulos2021gentle, shafer2008tutorial, balasubramanian2014conformal, vovk2022algorithmic} as a principled way to quantify uncertainty and support adaptive-latency decisions in SNNs. 

In the SpikeCP method introduced in \cite{chen2023spikecp}, the SNN produces \emph{set predictions} consisting of a subset of the set of all possible outputs. For instance, given as an input electroencephalography (EEG) or electrocardiography (ECG) time series, a set predictor determines a set of plausible conditions that a doctor may need to test for. Accordingly, for many applications, set predictors provide actionable information, while also offering an inherent measure of uncertainty in the form of the size of the predicted set \cite{angelopoulos2021gentle}. SpikeCP leverages the theoretical properties of CP to define reliable stopping rules based on the size of the predicted set.

\noindent \textbf{Motivation:} Predictive uncertainty can be decomposed into \emph{aleatoric uncertainty}, which refers to the inherent randomness of the data-generation mechanism, and \emph{epistemic uncertainty}, which arises due to the limited knowledge that can be extracted from a finite data set \cite{hullermeier2021aleatoric,simeone2022machine}. While aleatoric uncertainty is captured by individual machine learning models, like SNNs, epistemic uncertainty is typically accounted for by using \emph{ensembles} of models. In particular, epistemic uncertainty is quantified by gauging the level of \emph{disagreement} among the models in the ensembles \cite{hullermeier2021aleatoric,simeone2022machine}. 
By relying on conventional SNN models, SpikeCP does not attempt to quantify \emph{epistemic uncertainty}, focusing only on aleatoric uncertainty quantification.  The application of Bayesian learning and model \emph{ensembling} as means to quantify epistemic uncertainty in SNNs was investigated in \cite{skatchkovsky2022bayesian,katti2023bayesian,cai2018vibnn}, showing improvements in standard calibration metrics. 

In this paper, we propose to enhance the uncertainty quantification capabilities of SpikeCP by implementing ensemble SNN models for the purpose of improving the reliability of stopping decisions. Intuitively, an ensemble of multiple models can decide when to stop more reliably by selecting times at which most models agree that the current accuracy level is sufficient. The proposed method relies on tailored information pooling strategies across the models in the ensemble that preserve the theoretical guarantees of CP and SpikeCP.

\noindent \textbf{Main contributions:} The main contributions of this work are summarized as follows.\\
\noindent $\bullet$ We propose a novel ensemble-based SNN model that can reliably decide when to stop, producing set predictions with coverage guarantees and with an average latency that is significantly lower than the state of the art.\\
\noindent $\bullet$ We compare two ensembling stategies -- \emph{deep ensembles} (DE) \cite{lakshminarayanan2017simple, ganaie2022ensemble} and Bayesian learning via \emph{variational inference} (VI) \cite{simeone2022machine,skatchkovsky2022bayesian} -- and introduce two methods to efficiently combine the decisions from multiple models, namely \emph{confidence merging} (CM) and \emph{p-variable merging} (PM). In both cases, the resulting set predictors satisfy theoretical reliability guarantees.\\
 \noindent $\bullet$ Experiments show that VI-based ensembles with PM significantly reduce the average latency of state-of-the-art methods, while maintaining reliability guarantees.

\noindent \textbf{Organization:} The remainder of the paper is organized as follows. Section \ref{problem definition} presents the problem, and Section \ref{dcsnn} reviews the DC-SNN, while Section \ref{bspikecp} introduces the proposed framework. Section \ref{exp} describes the experimental setting and results. 

\begin{figure}[htp]
	\centering
	\includegraphics[width=6.1in]{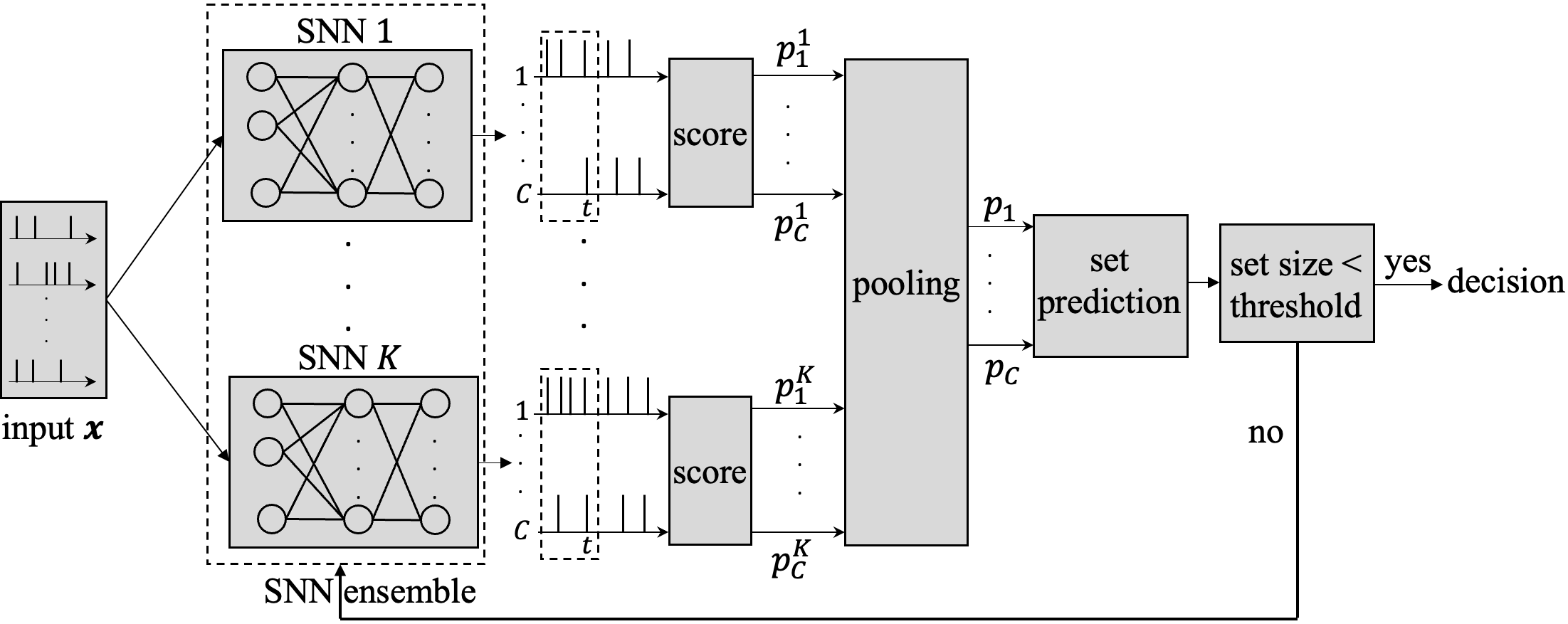}
	\caption{In the proposed system,  an ensemble of $K$ SNN models processes an input $\mv x$ agreeing on when to stop in order to make a classification decision. Each $k$th SNN model produces a score 
           $p_c^k$ for every candidate class $c=1,...,C$. The scores are combined to determine in an adaptive way whether to stop inference or to continue processing the input.}
	\label{gmodel}
\end{figure} 

\section{Problem Definition} \label{problem definition}
In this paper, we study adaptive-latency multi-class classification for time series via SNNs \cite{li2023seenn,li2023unleashing,  chen2023spikecp}. As illustrated in Fig.~1, unlike prior work \cite{li2023seenn,li2023unleashing,  chen2023spikecp}, we propose to enhance the reliability of stopping decisions by explicitly accounting for epistemic uncertainty when deciding whether to stop or to continue processing the input. The end goal is to  produce reliable set predictions with complexity and latency tailored to the difficulty of each example. In this section, we start by defining the problem and performance metrics.

\subsection{Multi-Class Classification with SNNs}
We wish to classify a vector time series $\mv x = \mv x_1,\mv x_2, ...$, with $N \times 1$ time samples $\mv x_t=[x_{t,1}, ..., x_{t,N}]$ into $C$ classes using an SNN model. The entries of input vector $\mv x_t$ can be arbitrary, although typical SNN implementations assume binary inputs  \cite{shrestha2023efficient}. As shown in Fig. 1, based on the time samples $\mv x^t=(\mv x_1, ..., \mv x_t)$ observed so far, at any time $t$, the $C$ read-out neurons of the SNN produce the $C \times 1$ binary vector $\mv y_t=[y_{t,1},..., y_{t,C}]$, with entries equal to 1 representing spikes. 

Internally, an SNN model can be viewed as a recurrent neural network (RNN) with binary activations.  Its operation is defined by a vector $\mv \theta$ of synaptic weights, which determines the response of each spiking neuron to incoming spikes. As in most existing art and implementations, we adopt a standard spike response model (SRM) \cite{gerstner2008spike} for the spiking neurons. 

Carrying out decision on the basis of the outputs of the  $C$ read-out neurons is typically achieved by \emph{rate decoding} \cite{10016643}. In rate decoding, at each time $t$, the SNN maintains a \emph{spike count vector} $\mv r(\mv x^t)=[r_1(\mv x^t), ..., r_C(\mv x^t)]$ in which each $c$th entry
\begin{align}
r_c(\mv x^t)= \sum_{t^{\prime}=1}^t y_{t^{\prime},c} \label{scount}
\end{align}
counts the number of spikes emitted so far by read-out neuron $c$. A normalized measure of 
\emph{confidence} can then be obtained via the softmax function as \cite{10016643}
\begin{align}
f_c(\mv x^t)=e^{r_c(\scalebox{0.7}{\mv x}^t)}/\sum_{c^{\prime}=1}^C e^{r_{c^{\prime}}(\scalebox{0.7}{\mv x}^t)}, \label{gprob}
\end{align}
for each class $c$. Conversely, the \emph{loss} assigned by the SNN model to label $c$ for input $x^t$ is given by the \emph{log-loss}  \begin{align}
    s_c(\mv x^t)=- \log f_c(\mv x^t). \label{nc}
 \end{align}The general goal of this work is to make reliable classification decisions at the earliest possible time $t$ on the basis of the confidence levels (\ref{gprob}), or equivalently of the losses (\ref{nc}), produced by SNN classifiers.

 \subsection{Ensemble Inference and Learning for SNNs} \label{ensemble}
Conventional SNN models consist of a single SNN making decisions on the basis of the confidence levels (\ref{gprob}), or \eqref{nc}, at a fixed time $t=T$. Neuroscience has long explored the connection between networks of spiking neurons and Bayesian reasoning \cite{doya2007bayesian}, and the recent work  \cite{skatchkovsky2022bayesian}
has  explored the advantages of Bayesian learning and model ensembling in terms of uncertainty quantification  for SNN classifiers. In this work, we leverage the enhanced uncertainty quantification capabilities of ensemble models to improve the reliability of adaptive-latency decision making via SNN models. 

As illustrated in Fig.~1,  in the considered setting, $K$ \emph{pre-trained} SNN classifiers are used in parallel on an input sequence $\mv x_1, \mv x_2,...$. The operation of each $k$th SNN classifier is defined by a vector $\mv \theta^k$ of synaptic weights as explained in the previous subsection. We specifically consider two design methods for the ensembles, namely \emph{deep ensembles} (DE) \cite{ganaie2022ensemble} and \emph{Bayesian learning} via \emph{variational inference} (VI)  \cite{simeone2022machine}.

In DE, the $K$ models are obtained by running conventional SNN training methods based on surrogate gradient \cite{neftci2019surrogate} with $K$ independent weight initializations, with each weight selected in an independent identically distribution (i.i.d.) manner as Gaussian $\mathcal{N}(0,\sigma^2)$ variables for some fixed variance $\sigma^2$. In contrast, in VI, assuming an i.i.d. Gaussian prior distribution $\mathcal{N}(0,\sigma^2)$ for the model parameter vector $\mv \theta$, one optimizes over a variational posterior distribution $\mathcal{N}(\mv \mu, \mv \zeta^2)$ parameterized by mean vector $\mv \mu$ and diagonal covariance matrix with diagonal elements given by vector $\mv \zeta^2$. The optimization is done by using gradient descent via the reparameterization trick \cite{skatchkovsky2022bayesian}. At inference time, the $K$ models are generated by sampling the weight vectors $\mv \theta^k$ from the optimized distribution $\mathcal{N}(\mv \mu, \mv \zeta^2)$. 

With DE, generating the $K$ models in the ensemble requires retraining from scratch, while this can be done by simply drawing Gaussian variables in the case of VI. Therefore, with DE, the ensemble should be practically shared across many input test sequences, while for VI it is possible to draw new ensembles more frequently, possibly even for each new input. 

\subsection{Set Prediction and Delay-Adaptivity}
As mentioned, we focus on delay-adaptive classifiers in which the time at which a decision is made is a function of the input $\mv x$ through the vector $\mv f(\mv x^t)=[f_1(\mv x^t),...,f_C(\mv x^t)]$ of confidence levels (\ref{gprob}) produced by the read-out neurons. Intuitively, when the model confidence is high enough, the classifier can produce a decision. We denote as $T_s(\mv x)$ the time at which a decision is made for input $\mv x$. Furthermore, we allow the decision to be in the form of a set $\Gamma(\mv x)\subseteq \{1,...,C\}$ of the set of $C$ labels \cite{angelopoulos2021gentle}. As mentioned in Sec. \ref{intro}, set decisions provide actionable information in many applications of interest, such as for robotics, medical diagnosis, and language modelling, and they provide a measure of uncertainty via the predicted set's size $|\Gamma(\mv x)|$ \cite{angelopoulos2021gentle}.

The performance of the classifier is measured  in terms of reliability and latency. A predictive set $\Gamma(\mv x)$ is said to be \emph{reliable} if the probability that the correct label $c$ is included in the set is no smaller than a pre-determined target accuracy $p_{\rm targ}$, i.e.,
\begin{align}
    \Pr(c \in \Gamma(\mv x)) \geq p_{\rm targ}, \label{reliability}
\end{align}
where the probability is taken with respect to the distribution of the test example $(\mv x, c)$,  as well as of the calibration data to be discussed next. The latency of the set prediction is defined as $\mathbb{E}[{T_s(\mv x)}]$, where the expectation is taken over the same distribution as for \eqref{reliability}.

The models are assumed to be pre-trained, and we assume to have access to a separate \emph{calibration data set} 
\begin{align}
    \mathcal{D}^{\rm cal}=\{(\mv x[i], c[i])\}_{i=1}^{|\mathcal{D}^{\rm cal}|}, 
\end{align} with $|\mathcal{D}^{\rm cal}|$ examples $(\mv x[i], c[i])$ generated i.i.d. from the same distribution followed by the test example $(\mv x, c)$ \cite{chen2023spikecp, angelopoulos2021gentle}. As we will discuss in the next section, calibration data is used to optimize the process of deciding when to stop so as to guarantee the reliability requirement (\ref{reliability}).

\section{Ensemble-based Adaptive Point Classification via SNNs} \label{dcsnn}
In this section, we first review dynamic-confidence SNN (DC-SNN), a point predictor for delay-adaptive SNN classification \cite{li2023unleashing}, and then introduce the ensemble-based version.

\subsection{DC-SNN} 

DC-SNN produces a decision at the first time $t$ for which the maximum confidence level across all possible classes is larger than a fixed \emph{target confidence level} $p_{\textrm{th}}\in(0,1)$. Accordingly, the stopping time is given by 
\begin{align}
T_s(\mv x)=&\min_{t \in \{1,...,T\}} t  ~\text{ s.t.}~~ \max_{c\in\mathcal{C}} f_{c}(\mv x^t)\geq p_{\textrm{th}} \label{stop}   
\end{align}  
if there is a time $t<T$ that satisfies the constraint; and $T_s(\mv x)=T$ otherwise. The rationale for this approach is that, by (\ref{stop}), if $T_s(\mv x)<T$, the classifier has a confidence level no smaller than $p_{\textrm{th}}$ on the decision 
\begin{equation}
    \hat{c}(\mv x)=\arg\max_{c\in\mathcal{C}} f_{c}(\mv x^{T_s(\scalebox{0.7}{\mv x})}). \label{decision}
\end{equation} 

If the SNN classifier is \emph{well calibrated}, the confidence level coincides with the true accuracy of the decision given by the class $\argmax_{c\in \mathcal{C}} f_c (\mv x^t)$ at all times $t$. Therefore, setting the target confidence level $p_{\text{th}}$ to be equal to the target accuracy $p_\text{targ}$, i.e., $p_\text{th} = p_\text{targ}$, guarantees a zero, or negative, reliability gap for the adaptive decision (\ref{decision}) when $T_s(\mv x)<T$. However, the assumption of calibration is typically not valid.  To address this problem, reference  \cite{li2023unleashing} introduced a solution based on the use of a calibration data set.

Specifically, DC-SNN evaluates the empirical accuracy of the decision (\ref{decision}), i.e., \begin{equation}\hat{\mathcal{A}}^{\textrm{cal}}(p_\mathrm{th})=\frac{1}{|\mathcal{D}^{\rm cal}|}\sum_{i=1}^{|\mathcal{D}^{\rm cal}|}\mathbbm{1}(\hat{c}(\mv x[i])=c[i]),\end{equation} where $\mathbbm{1}(\cdot)$ is the indicator function,  for a grid of possible values of  the target confidence level $p_{\textrm{th}}$. Then, it  chooses the minimum value $p_{\textrm{th}}$ that ensures the inequality $\hat{\mathcal{A}}^{\textrm{cal}}(p_\mathrm{th})\geq p_\text{targ}$, so that the calibration accuracy exceeds the target accuracy level $p_\text{targ}$; or the smallest value $p_{\rm th}$ that maximizes $\hat{\mathcal{A}}^{\textrm{cal}}(p_{\rm th})$ if the constraint  $\hat{\mathcal{A}}^{\textrm{cal}}(p_\mathrm{th})\geq p_\text{targ}$ cannot be met.

\subsection{Ensemble-based DC-SNN}
Following Sec. \ref{ensemble}, one can directly extend DC-SNN to implement approximate Bayesian learning by means of VI and DE methods. Accordingly, at inference time, a decision is made on the basis of $K$ SNN models from a trained ensemble, which is fixed in the case of DE and randomly generated for VI. In this subsection, we briefly describe the decision procedure for a Bayesian version of DC-SNN.

Given some input $\mv x$, each $k$th model produces a confidence value $f^k_c(\mv x^t)$ for the pair $(\mv x^t, c)$. Implementing standard Bayesian model averaging, the confidence values $f_c^k(\mv x^t)$, $k=1, \ldots, K$, for all models are then pooled by averaging as \begin{align}
    f_c(\mv x^t)=\frac{1}{K} \sum_{k=1}^K f^k_c(\mv x^t). \label{enpro}
\end{align}
The ensemble probability $f_c(\mv x^t)$ in \eqref{enpro} is finally applied in \eqref{stop} and \eqref{decision} to obtain the final decision.

\section{Ensemble-based Adaptive Set Classification via SNNs} \label{bspikecp}
In this section, we introduce \emph{ensemble-based SpikeCP}, a novel framework for delay-adaptive classification that wraps around any pre-trained ensemble of SNN classifiers, including ensembles obtained via DE and VI. We propose two implementations corresponding to different ways of pooling information across the $K$ models in the ensemble.

\subsection{SpikeCP}
We first review SpikeCP \cite{chen2023spikecp}, which applies to a single SNN model, i.e., with $K=1$. The presentation here, unlike in \cite{chen2023spikecp}, adopts the language of p-variables (see, e.g., \cite{papadopoulos2008inductive, vovk2022algorithmic}) in order to facilitate the extension to ensemble models. 

SpikeCP fixes a pre-determined set of \emph{checkpoint times} $\mathcal{T}_s\subseteq \{1,...,T\}$ at which inference may stop to produce a decision. The information available to determine whether to stop or not are the losses $\{s_c(\mv x^t)\}_{c=1}^C$ in (\ref{nc}) for the current input $\mv x^t$, as well as the corresponding  losses $s_{c[i]}(\mv x^t[i])$  for the calibration data points indexed by  $i=1,...,|\mathcal{D}^\mathrm{cal}|$. For each class $c$, SpikeCP computes the quantity
\begin{align}
    \label{eq:conformal_p_value}
    { p_c}(\mv x^t) = \frac{ \sum_{i=1}^{|\mathcal{D}^\text{cal}|}\mathbbm{1}(s_{c}(\mv x^t) \leq s_{c[i]}(\mv x^t[i]) )+1}{|\mathcal{D}^\text{cal}|+1}, 
\end{align} 
where $\mathbbm{1}(\cdot)$ equals 1 if the argument is true and 0 otherwise.  The quantity (\ref{eq:conformal_p_value})  corresponds, approximately, to the fraction of calibration data points whose loss is no smaller than the loss for label $c$ when assigned to the current test input $\mv x^t$. The corrections by 1 at numerator and denominator are required to guarantee the following property, which follows from the standard theory of CP \cite[Proposition 1]{vovk2012conditional}. 

\begin{theorem}
    Let $\mathcal{D}^{t, \rm cal}=\{(\mv x^t[i], c[i])\}_{i=1}^{|\mathcal{D}^{\rm cal}|}$ be the calibration data set with samples up to time $t$, and define as $\mathcal{H}_c^t$ the hypothesis that the pair $(\boldsymbol{x}^t, c)$ and the calibration data $\mathcal{D}^{t, \rm cal}$ are i.i.d. The quantity \eqref{eq:conformal_p_value} is a p-variable for null hypothesis $\mathcal{H}_c^t$, i.e., we have the conditional probability
\begin{align}
\Pr(p_c(\boldsymbol{x}^t) \leq  \alpha|\mathcal{H}^t_c)\leq \alpha, \label{valid}
\end{align}
for all $\alpha \in (0,1)$, where the probability is taken over the distribution of test and calibration data.
\end{theorem}

At each checkpoint $t\in \mathcal{T}_s$, SpikeCP constructs a predictive set by including all classes $c$ with p-variable larger than threshold $\alpha$
\begin{align}
\Gamma(\boldsymbol{x}^t)=\{c\in\mathcal{C}: {p}_c(\boldsymbol{x}^t) > \alpha\}. \label{pset}
\end{align} By (\ref{valid}), the probability that the set  (\ref{pset}) does not include the true test label $c$ is smaller or equal than $\alpha$, or equivalently \cite[Proposition 1]{vovk2012conditional} 
\begin{equation}\label{eq:guarantee}\Pr(c \in \Gamma(\mv x^t)) \geq 1-\alpha.\end{equation}  
Accordingly, SpikeCP sets $\alpha=(1-p_{\rm targ})/|\mathcal{T}_s|$ to ensure that condition (\ref{eq:guarantee}) is satisfied irrespective of which checkpoint is selected. As detailed in \cite{chen2023spikecp}, this is a form of \emph{Bonferroni correction} \cite[Appendix 2]{hochberg1987multiple}. 

SpikeCP stops inference at the first time $T_s(\mv x)$ for which the size of the predicted set is smaller than a target set size $I_{\rm th}$, so the stopping time is given by
\begin{align}
    T_s(\mv x) = \min\{{t\in \mathcal{T}_s}: |\Gamma(\mv x^t)|\leq I_{\rm th} \}.
\end{align} The threshold $I_{\rm th}$ is a design choice that is dictated by the desired informativeness of the resulting set predictor. For any threshold $I_{\rm th}$, by construction, SpikeCP satisfies the reliability property (\ref{reliability}) \cite[Theorem 1]{chen2023spikecp}.

\subsection{Ensemble-based SpikeCP with Confidence Merging}
In the proposed ensemble-SNN architecture in Fig. 1, each SNN classifier parameterized by $\mv \theta^k$, $k=1,...,K$, produces a generally different probability $f^k_c(\mv x^t)$ in \eqref{gprob}, or correspondingly a different loss $s^k_c(\mv x^t)$,  for each class $c$ given an input $\mv x^t$. In this paper, we study and compare two combining mechanisms.

First, in order to produce a confidence level for each possible label $c$, the confidence levels output by the $K$ models in the ensemble can be combined using the generalized mean \cite{koliander2022fusion}
\begin{align}
    f_c(\mv x^t)=\bigg(\frac{1}{K} \sum_{k=1}^K \big(f^k_c(\mv x^t)\big)^r\bigg)^{1/r} \label{mgcount}
\end{align}
for some integer $r\in[-\infty,+\infty]$. When $r=1$, the ensemble probability (\ref{mgcount}) reduces to standard model averaging. Other values of $r$ may in practice be advantageous, e.g., to enhance robustness \cite{oh2016generalized, gou2019generalized}, with maximum operation recovered for $r=\infty$ and the minimum operation obtained with $r=-\infty$. The probability \eqref{mgcount} is used to calculate the score via \eqref{nc}, which is then directly used in \eqref{eq:conformal_p_value} and \eqref{pset} to determine the set predictor. Note that the same combination in (\ref{mgcount}) is also applied to calibration data. By the same arguments as for SpikeCP, this approach guarantees the reliability condition (\ref{reliability}) by setting {$\alpha=(1-p_{\rm targ})/|\mathcal{T}_s|$.

\subsection{Ensemble-based SpikeCP with P-Variable Merging}
Given the reliance of the predicted set \eqref{pset} on p-variables, merging directly the confidence levels may be suboptimal \cite{meng1994posterior}. Accordingly, in this subsection, we explore the idea of pooling directly the p-variables, rather than combining confidence levels. To this end,  we first calculate the losses for the calibration set by using the $k$th model as $\{s^k_{c[i]}(\mv x^t[i])\}_{i=1}^{|\mathcal{D}^{t,\rm{cal}}|}$ for $k=1,...,K$. Then, for a test input $\mv x^t$, we evaluate the p-variable \eqref{eq:conformal_p_value} for the $k$th model as 
\begin{align}
    { p}_c^k(\mv x^t) = \frac{ 1+\sum_{i=1}^{|\mathcal{D}^\text{cal}|}\mathbbm{1}(s^k_{c}(\mv x^t) \leq s^k_{c[i]}(\mv x^t[i]) )}{|\mathcal{D}^\text{cal}|+1}. 
\end{align}
The p-variables $\{p_c^k(\mv x^t)\}_{k=1}^K$ are then pooled by  using any \emph{p-merging} function $F(\cdot)$, as defined next.
\begin{definition}[\cite{vovk2022admissible, vovk2020combining}]
A function $F: [0, 1]^K \rightarrow [0,\infty)$ is said to be a p-merging function if, when the inputs are p-variables, the output is also a p-variable, i.e., we have the inequality
\begin{align}
\label{eq:p-merging}
\Pr(F\big({ p_c}^1(\boldsymbol{x}^t), ..., { p_c}^K(\boldsymbol{x}^t)\big)\leq \alpha') \leq \alpha',  \text{ for all } \alpha' \in (0,1),
\end{align}
where the probability is taken over the joint distribution of the $K$ input p-variables.
\end{definition}

Using the merged p-value generated as
\begin{align}\label{eq:p-merging_for_BSpikeCP}
    p_c(\mv x^t) = F\big({p_c}^1(\mv x^t), ..., { p_c}^K(\mv x^t)\big)
\end{align} for any p-merging function $F(\cdot)$, 
the predictive set can be constructed by following \eqref{pset}. By definition of p-merging function, the resulting set predictor also satisfies the reliability condition (\ref{reliability}).

In the experiments reported in the next section, we focus on the class of p-merging functions of the form \cite{vovk2020combining}
\begin{align}
F(p^1,...,p^K) = a_{r} \bigg(\frac{1}{K} \sum_{k=1}^K \big({ p}^k\big)^r\bigg)^{1/r}, \label{emerging}
\end{align} 
    where  $a_{r}$ is a constant chosen so as to ensure (\ref{eq:p-merging}) as specified in \cite[Table 1]{vovk2020combining}.  For example, setting $r=-\infty$, and correspondingly $a_r=K$, yields the p-merging function $F(p^1,...,p^K)  = K \min(p^1,...,p^K)$, while setting $r=\infty$ with $a_{\infty}=1$  yields $F(p^1,...,p^K)  = \max(p^1,...,p^K)$.

\section{Experiments} \label{exp}

For numerical evaluations, we consider the standard DVS128 Gesture dataset \cite{amir2017low}, MINIST-DVS dataset \cite{serrano2015poker} and the CIFAR-10 dataset. The first dataset represents a video recognition task, and the latter two represent image classification tasks. The calibration data set $\mathcal{D}^\text{cal}$ is obtained by randomly sampling $|\mathcal{D}^\text{cal}|=50$ examples from the test set, with the rest used for training, which is done via the  surrogate gradient method \citep{neftci2019surrogate}. The length of the time series is $T=80$ samples, and we fix the set of possible checkpoints as $\mathcal{T}_s = \{20, 40, 60, 80\}$, and the target set size to $I_{\rm th}=3$. The target accuracy $p_{\rm targ}$ is set to $0.9$.

We compare the performance of ensemble-based SpikeCP using DE or VI equipped with confidence merging (CM) or p-variable merging (PM) and ensemble-based DC-SNN. For DE, we follow standard random initialization made available by PyTorch, while for VI we set the prior distribution to have variance 0.03. The parameter $r$ in \eqref{mgcount} for CM is set to $1$, yielding standard model averaging \cite{skatchkovsky2022bayesian}; while $r$ in \eqref{emerging} for PM is set to $r=45$, with $a_r = K^{1/r}$ following \cite[Table 1]{vovk2020combining}, based on the numerical minimization of latency on a held-out data set. The results are averaged over $50$ different realizations of calibration and test data sets, and the number of ensemble $K$ is set to 6. For fair comparison, we apply the stopping rule defined in Sec. \ref{dcsnn} to obtain the stopping time, and use a top-3 predictor to produce a set $\Gamma^d(\mv x)$ for ensemble-based DC-SNN.

\subsection{MNIST-DVS Dataset}
The MNIST-DVS dataset contains time series recorded from a DVS camera that is shown moving handwritten digits from $``0"$ to $``9"$ on a screen.  The data set contains $8,000$ training examples, as well as $2,000$ examples used for calibration and testing. For this experiment, we adopt a fully connected SNN with one hidden layer having $1,000$ neurons. 

\begin{figure}[htp]
	\begin{center}
		\includegraphics[width=3.2in]{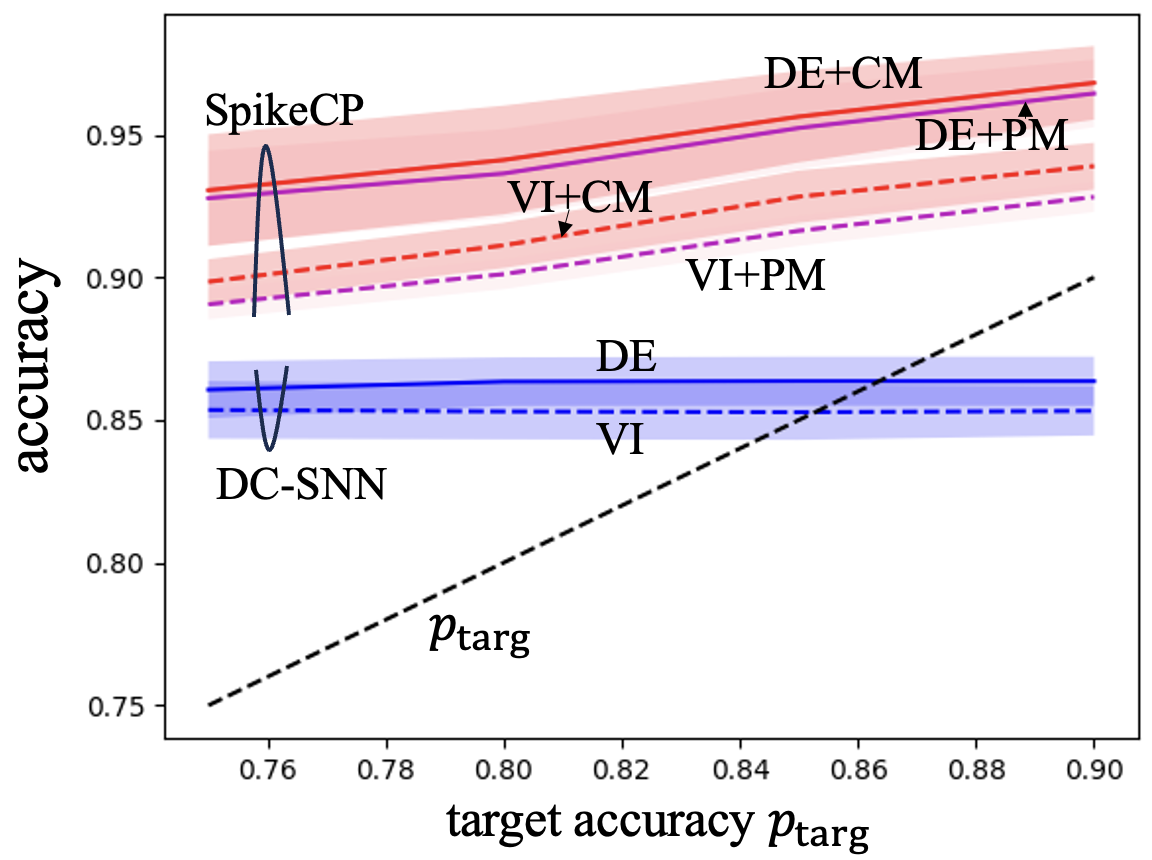}\hfill
		\includegraphics[width=3.2in]{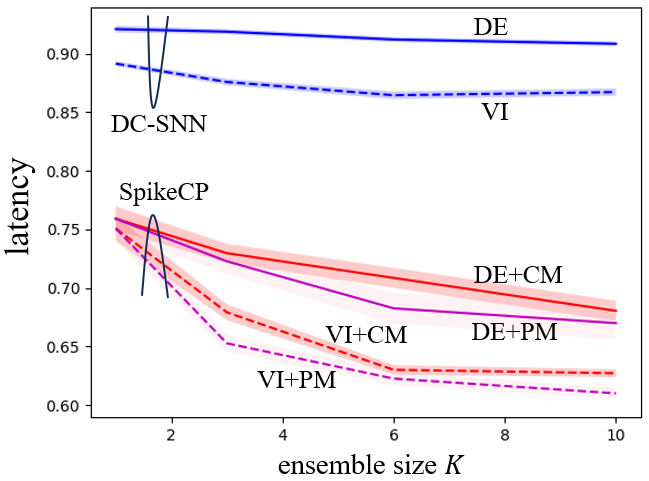}
	\end{center}
        \vspace{-0.2cm}
	\caption{Accuracy ($\Pr(c \in \Gamma^d(\mv x))$ for ensemble-based DC-SNN and $\Pr(c \in \Gamma(\mv x))$ for ensemble-based SpikeCP) and normalized latency $\mathbb{E}[T_s(\mv x)]/T$ as a function of the target accuracy $p_{\rm targ}$ for MNIST-DVS dataset.}
\vspace{-0.2cm}
\label{r1}
\end{figure}

Fig.~\ref{r1} reports accuracy -- $\Pr(c \in \Gamma^d(\mv x))$ for ensemble-based DC-SNN and $\Pr(c \in \Gamma(\mv x))$ for ensemble-based SpikeCP -- and normalized latency $\mathbb{E}[T_s(\mv x)]/T$  as a function of the target accuracy $p_\mathrm{targ}$. Ensemble-based DC-SNN increases the decision latency as the target probability $p_{\mathrm{targ}}$ increases, in order to meet the reliability condition. However, the reliable decision is only attained by DC-SNN when $p_{\rm targ}$ is small.  In contrast, ensemble-based SpikeCP is always reliable, irrespective of the target accuracy. Furthermore, ensemble-based SpikeCP using VI and PM requires smaller latency to achieve the target accuracy.

\begin{figure}[t!]
	\begin{center}
		\includegraphics[width=3.2in]{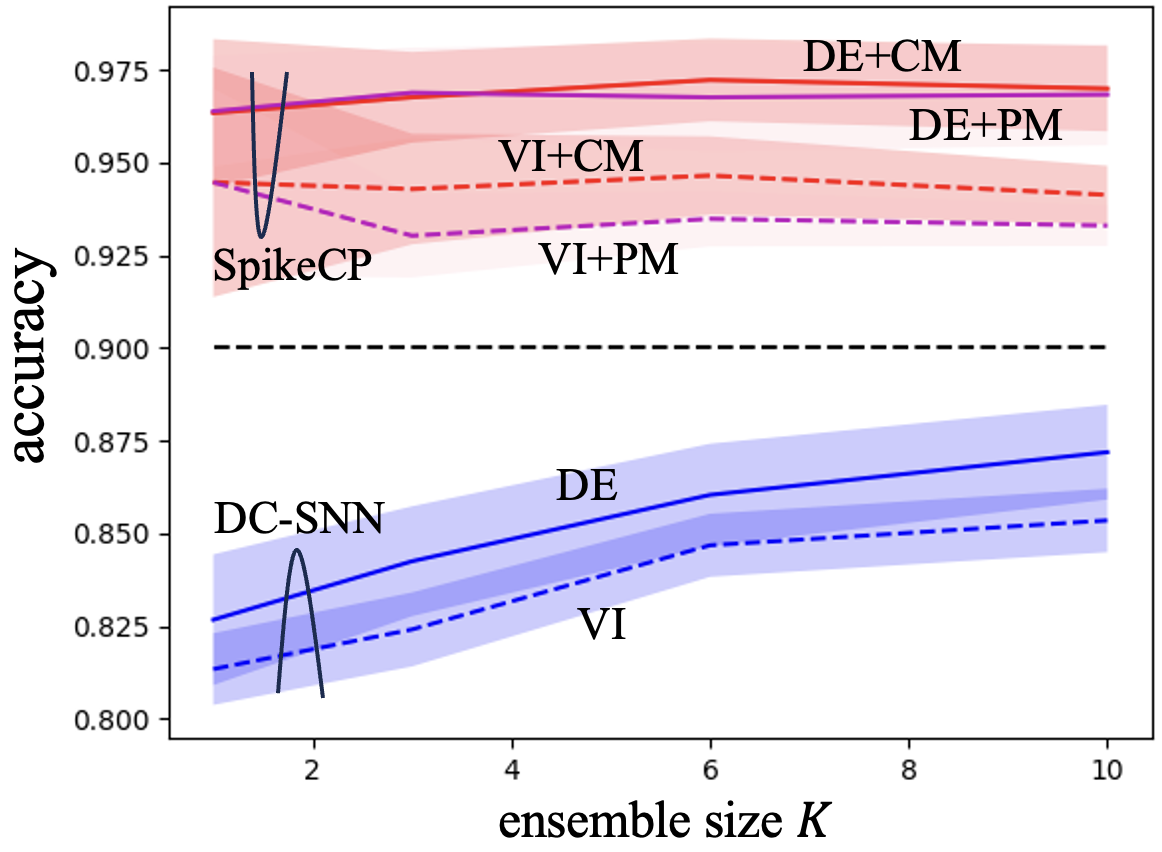}\hfill
		\includegraphics[width=3.2in]{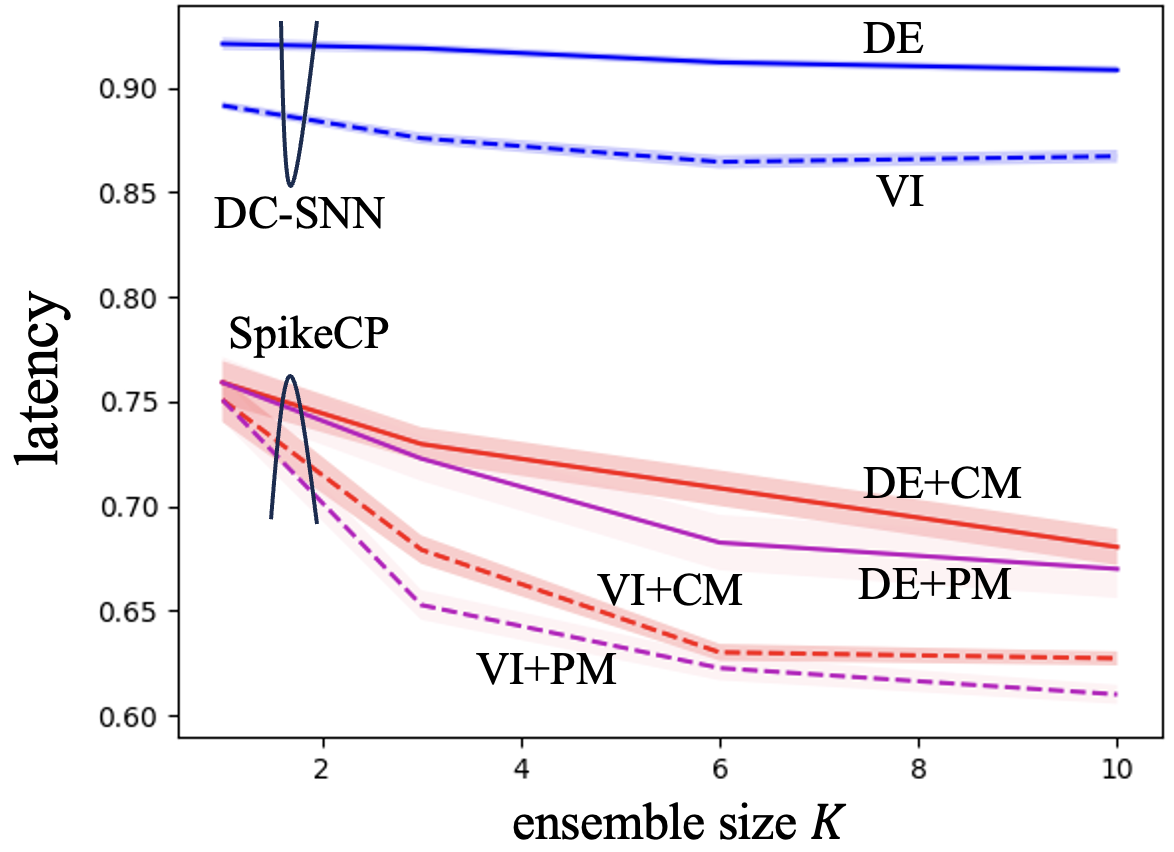}
	\end{center}
        \vspace{-0.2cm}
	\caption{Accuracy ($\Pr(c \in \Gamma^d(\mv x))$ for ensemble-based DC-SNN and $\Pr(c \in \Gamma(\mv x))$ for ensemble-based SpikeCP) and normalized latency $\mathbb{E}[T_s(\mv x)]/T$ as a function of the ensemble size $K$ for MNIST-DVS dataset.}
\vspace{-0.2cm}
\label{r2}
\end{figure}

In Fig.~\ref{r2}, we show the accuracy and normalized latency as a function of the ensemble size. With a larger ensemble size, both ensemble-based DC-SNN and SpikeCP exhibit reduced latency in reaching a final decision. While SpikeCP maintains its reliability guarantee, DC-SNN falls short of achieving the target accuracy. 

\begin{figure}[t!]
	\begin{center}
		\includegraphics[width=3.2in]{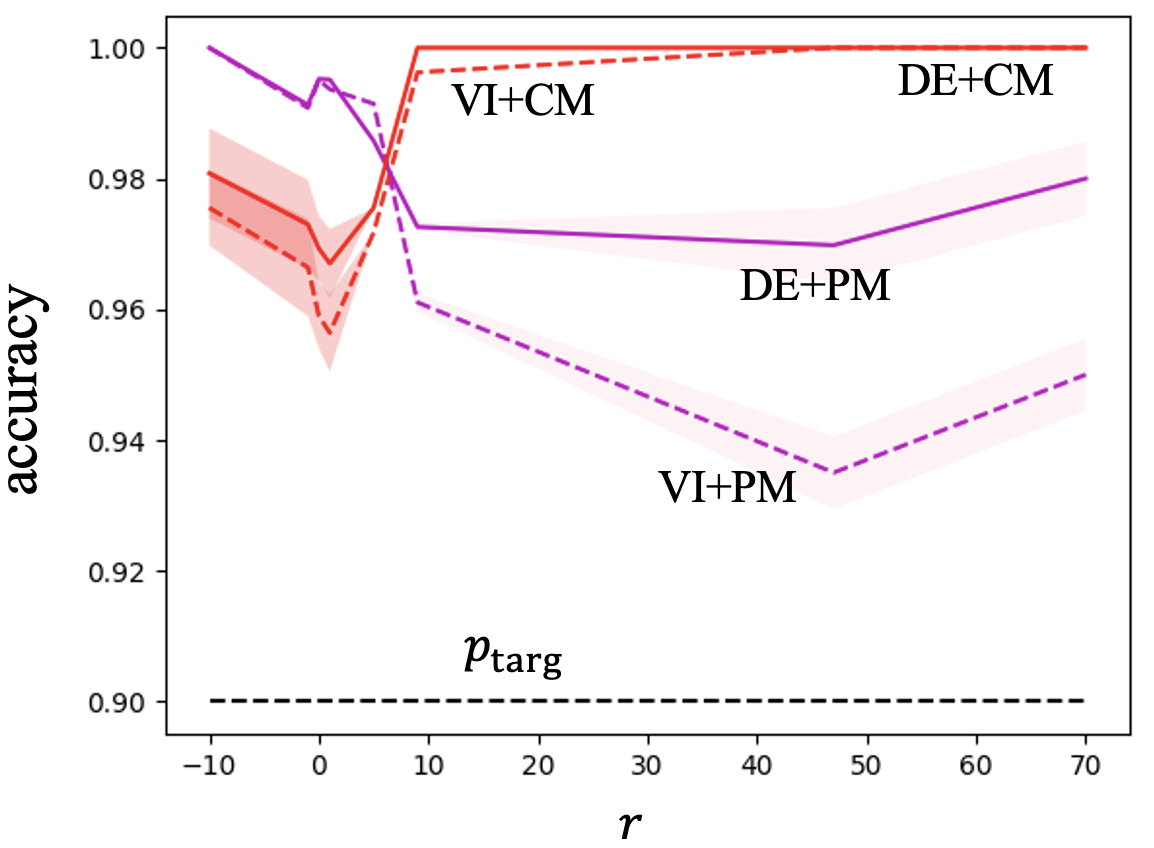}\hfill
		\includegraphics[width=3.2in]{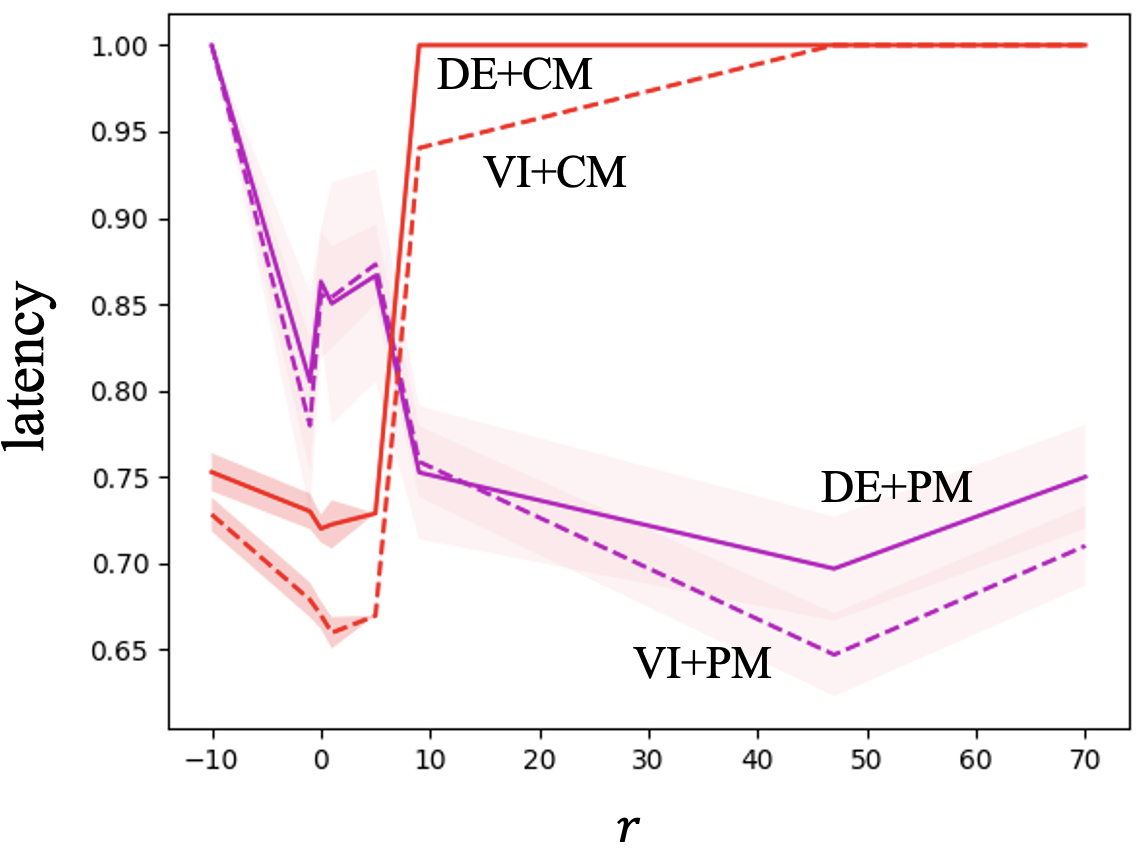}
	\end{center}
        \vspace{-0.2cm}
	\caption{Accuracy  $\Pr(c\in\Gamma(\mv x))$  and normalized latency $\mathbb{E}[T_s(\mv x)]/T$ as a function of the hyperparameter $r$ (in \eqref{mgcount} for SpikeCP with CM and in \eqref{emerging} for SpikeCP with PM) for MINIST-DVS dataset.}
\vspace{-0.2cm}
\label{r3}
\end{figure}

To explore the impact of the hyper-parameter $r$ in \eqref{mgcount} and \eqref{emerging} for ensemble-based SpikeCP, we show in Fig.~\ref{r3} the accuracy and normalized latency as a function of $r$. To ensure the p-merging function in \eqref{emerging} produces a valid p-value, we adopt different p-merging function $F(p^1, \ldots, p^K)$ under different $r$ as in \cite[Table 1]{vovk2020combining}.  CM pooling methods exhibit the lowest latency when $r$ is approximately around 1, aligning with the reduction to standard Bayesian learning, while PM  demonstrates a smaller latency with larger values of $r$.

\subsection{DVS128 Gesture Dataset}
The DVS128 Gesture data set collects videos from a DVS camera that is shown an actor performing one of $11$ different gestures under three different illumination conditions. We divide each time series into $T=80$ time intervals, integrating the discrete samples within each interval to obtain a (continuous-valued) time sample \citep{fang2021incorporating}. The dataset contains 1176 training data and 288 test data, from which 50 examples are chosen to serve as calibration data. The SNN architecture is constructed using a convolutional layer, encompassing batch normalization and max-pooling layer,  as well as a fully-connected layer as described in \citep{fang2021incorporating}. 

In Fig.~\ref{r4}, we show the accuracy, given by the probability $\Pr(c \in \Gamma(\mv x))$ in \eqref{reliability} and the average decision latency as a function of the ensemble size $K$ on DVS128 Gesture dataset. The performance of ensemble-based DC-SNN is similar to that on MNIST-DVS dataset, failing to meet the target accuracy. To highlight the performance of ensemble-based SpikeCP, we omit the performance of DC-SNN here. Confirming their theoretical properties, all ensemble-based SpikeCP schemes  meet the target accuracy $p_{\text{targ}}=0.9$. Furthermore, the average latency decreases with the ensemble size $K$, providing substantial improvements as compared to the original SpikeCP scheme with $K=1$ \cite{chen2023spikecp}. 

VI methods tend to  have a better performance in terms of latency, showcasing the benefits of VI as a more principled approach for Bayesian learning. Finally, PM generally yields  smaller latency values as compared to CM, indicating that merging p-variables offers a more efficient information pooling strategy.

\begin{figure}[t!]
	\begin{center}
		\includegraphics[width=3.2in]{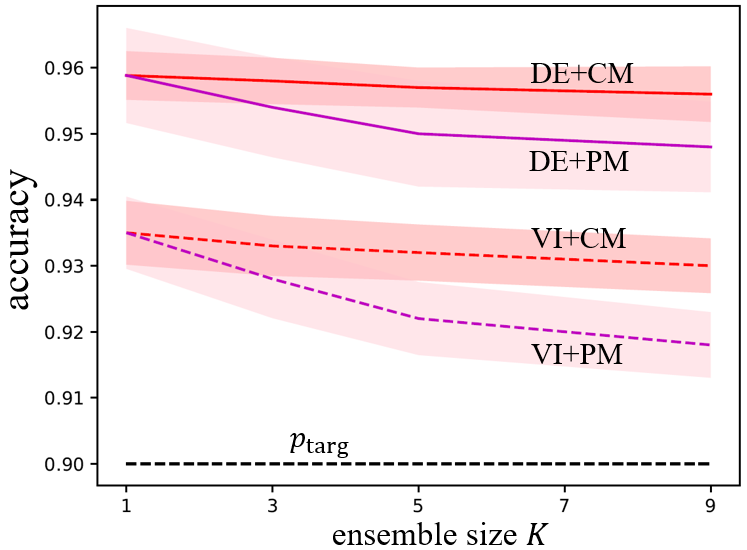}\hfill
		\includegraphics[width=3.2in]{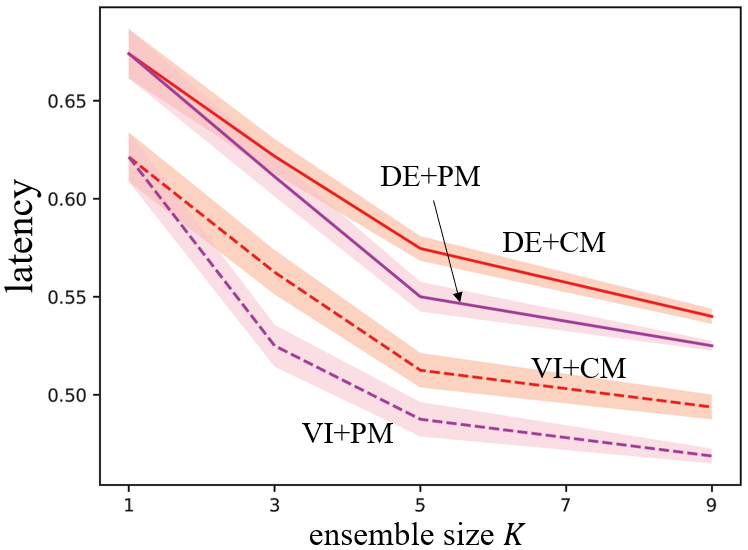}
	\end{center}
        \vspace{-0.2cm}
	\caption{Accuracy  $\Pr(c\in\Gamma(\mv x))$  and normalized latency $\mathbb{E}[T_s(\mv x)]/T$ as a function of the ensemble size $K$ for DVS128 Gesture dataset.}
\vspace{-0.2cm}
\label{r4}
\end{figure}

\subsection{CIFAR-10 Dataset}
The CIFAR-10 dataset consists of 60,000 $32 \times 32$ color images that are divided into 10 classes, with 6000 images per class. There are 50,000 training images and 10,000 test images.  We use $|\mathcal{D}^{\rm cal}|=50$ calibration samples, which are obtained by randomly selecting 50 data points from the test set. We adopt a ResNet-18 architecture in which conventional neurons are replaced with SRM neurons \citep{fang2021incorporating}. Each example is repeatedly presented to the SNN for $T=80$ times.

\begin{figure}[t!]
	\begin{center}
		\includegraphics[width=3.2in]{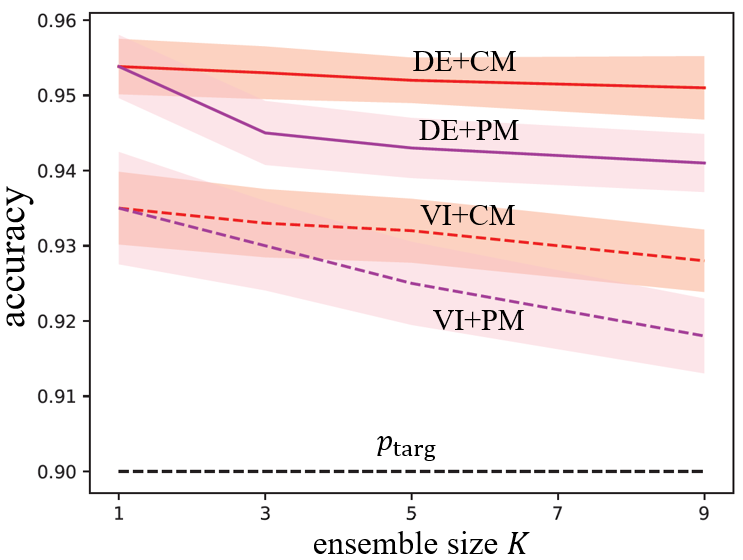}\hfill
		\includegraphics[width=3.2in]{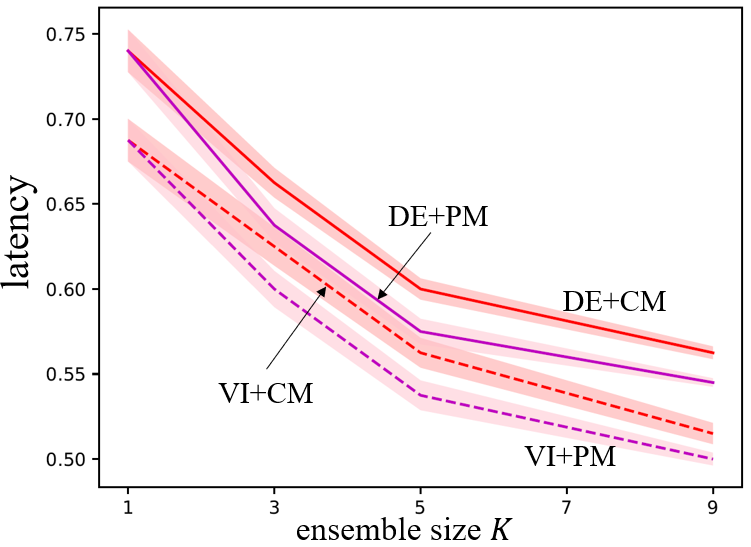}
	\end{center}
        \vspace{-0.2cm}
	\caption{Accuracy  $\Pr(c\in\Gamma(\mv x))$  and normalized latency $\mathbb{E}[T_s(\mv x)]/T$ as a function of the ensemble size $K$ for CIFAR-10 dataset.}
\vspace{-0.2cm}
\label{r5}
\end{figure}

In Fig.~\ref{r5}, we show the accuracy $\Pr(c\in\Gamma(\mv x))$  and normalized latency $\mathbb{E}[T_s(\mv x)]/T$ as a function of the ensemble size $K$ on CIFAR-10 dataset for ensemble-based SpikeCP. As per our theory, SpikeCP can guarantee the reliability condition  with all information pooling schemes. Furthermore,  VI with PM produces the best performance in terms of  latency.

\section{Conclusions}
In this work, we have introduced  ensemble-based SpikeCP, a novel delay-adaptive SNN set predictor with provable reliability guarantees. Ensemble-based SpikeCP  improves the epistemic uncertainty quantification capacity of SNNs, which, in turn,  enhances the reliability of stopping decisions for adaptive-latency classification. Intuitively, considering the predictions of multiple models supports the determination of a more reliable stopping time by focusing on time instants at which most models agree that the current accuracy level is sufficient. Our proposed approach relies on information pooling from ensemble models and provides theoretical guarantee of reliability.

Future work may consider applications of the proposed method to domains such as wireless communications, in which reliability and latency are essential performance criterion \cite{10016643}.

\small{
\bibliographystyle{IEEEtran}
\bibliography{references}}
\end{document}